\documentclass{article}
\usepackage{spconf,amsmath,graphicx,color}
\usepackage{multirow}
\usepackage{booktabs}
\usepackage{cite}

\def \eg {\emph{e.g.}}
\def \ie {\emph{i.e.}}

\title{Face Recognition on Point Cloud with cGAN-TOP for Denoising}
\name{Junyu Liu$^{\star}$ \qquad Jianfeng Ren$^{\star \dagger}$ \qquad Hongliang Sun$^{\star}$ \qquad Xudong Jiang $^{\ddagger}$\thanks{This work was supported in part by National Natural Science Foundation of China under Grant 72071116 and 71901115, and in part by Ningbo Municipal Bureau Science and Technology under Grant 2019B10026 and 2022Z173. Corresponding author: jianfeng.ren@nottingham.edu.cn.}}
\address{$^{\star}$School of Computer Science, University of Nottingham Ningbo China, China \\
$^{\dagger}$Nottingham Ningbo China Beacons of Excellence Research and Innovation Institute, \\
University of Nottingham Ningbo China, China\\
$^{\ddagger}$School of Electrical \& Electronic Engineering, Nanyang Technological University}

\begin{document}
%\ninept
%-------------------- Abstract --------------------%
\maketitle
\begin{abstract}
Face recognition using 3D point clouds is gaining growing interest, while raw point clouds often contain a significant amount of noise due to imperfect sensors. In this paper, an end-to-end 3D face recognition on a noisy point cloud is proposed, which synergistically integrates the denoising and recognition modules. Specifically, a Conditional Generative Adversarial Network on Three Orthogonal Planes (cGAN-TOP) is designed to effectively remove the noise in the point cloud, and recover the underlying features for subsequent recognition. A Linked Dynamic Graph Convolutional Neural Network (LDGCNN) is then adapted to recognize faces from the processed point cloud, which hierarchically links both the local point features and neighboring features of multiple scales. The proposed method is validated on the Bosphorus dataset. It significantly improves the recognition accuracy under all noise settings, with a maximum gain of $14.81\%$.
\end{abstract}

%-------------------- Key Words --------------------%

%-------------------- Introduction --------------------%
\section{Introduction}
\label{sec:intro}

% CVPR18/Learn-Large-scale-3D-Face,IJCB21/PointFace,,Hand-craft-2,Hand-craft-3
3D face recognition becomes increasing popular, as 3D data provide more information and are less affected by illumination \cite{CVPR18/Learn-Large-scale-3D-Face,IJCB21/PointFace,FeatureBoost-3D,Vessel-seg,HCI22/Face2Statistics}. Among all 3D models, point clouds are most commonly used \cite{arxiv/LDGCNN}. Early methods for 3D face recognition often manually specify geometric properties for feature extraction \cite{Hand-craft-1}. 2D deep learning methods \cite{2D-based-1,2D-based-2,2D-based-3} project point clouds as depth maps and use 2D networks for recognition, which do not fully utilize the 3D information \cite{arxiv/DGCNN}. 3D geometric-based methods directly process the sparse point sets using Graph Convolutional Networks \cite{IJCNN16/PointNet}. Most recent 3D face recognition models \cite{IJCB21/PointFace,3D-based,CVPR18/Learn-Large-scale-3D-Face} utilize PointNet \cite{IJCNN16/PointNet} or PointNet++ \cite{NIPS17/PointNet++}, in which sparse convolution kernels are directly applied on each point. But these methods overemphasize the point features, and may not fully utilize the global information. Furthermore, most of them assume clean 3D face point clouds, limiting their practical usability.

To handle the noisy raw data in 3D face recognition, researchers attempted to directly learn a model from noisy data \cite{AAAI21/Learn-low-quality,CVPR19/Led3D}, but these models can't flexibly handle point clouds with different types/levels of noise. Graph Convolutional networks \cite{CVPR19/3D-Point-Set-Upsampling,IJCNN16/PointNet,NIPS17/PointNet++} have been utilized to denoise regular objects, \eg, chairs, cars, and bottles, but they perform poorly on complex 3D face point clouds. Clean face models are usually thin surface-like models with one-point thickness, but the noisy face point clouds may drastically disturb the surface, challenging the design of networks. More importantly, we observe that most denoising methods only focus on producing visually clean point clouds and most recognition methods use mostly clean point clouds, leaving a loophole of an end-to-end face recognition method on the raw point clouds.
 
To tackle these challenges, an end-to-end framework for face recognition with denoising on 3D point clouds is proposed in this paper. Specifically, a Conditional Generative Adversarial Network on Three Orthogonal Planes (cGAN-TOP) is designed to reduce the noise in the point cloud, where in each plane, one of the 3D coordinates is treated as the feature (similar to the intensity level in an image) and the other two are kept as the coordinates. A cGAN \cite{yang2022rain} is designed for each plane to remove the noise in the feature (one of the coordinates) with the help of the other two coordinates. The three denoised coordinates are then combined to form the denoised point cloud. Finally, a Linked Dynamic Graph Convolutional Neural Network (LDGCNN) is adapted to comprehensively utilize both local and global point features for recognition.

Our contributions are three-fold: 1) An end-to-end model for face recognition on noisy point clouds is proposed, which integrates the cGAN-TOP denoising model and the LDGCNN recognition model. 2) The proposed cGAN-TOP removes the noise in the point cloud targeted at a better subsequent recognition. 3) The proposed method significantly outperforms state-of-the-art models under challenging conditions on the Bosphorus dataset \cite{Bosphorus}, with a maximum gain of $14.81\%$.

\vspace{-1em}
%-------------------- Proposed Methods --------------------%
\section{Proposed Method}
\label{sec:methods}
\subsection{Overview of Proposed Method}
Point clouds are sparse and unstructured \cite{IJCNN16/PointNet} (unlike densely structured models such as 2D images), enabling them to represent complex models flexibly. But recognition using 3D point clouds faces two main challenges. 1) The sparsity poses a challenge to point cloud processing since traditional 2D and 3D CNNs can only process structured models like images and volumetric models \cite{arxiv/LDGCNN}. 2) Raw point clouds often contain noise due to imperfect sensors \cite{CVPR19/Led3D}, which is significantly different from the noise in images, depth maps, or other volumetric models. The 3D coordinates of a point cloud serve as both features and spatial coordinates, while other modalities have a distinct difference between features and spatial coordinates, \eg, the spatial coordinates are commonly assumed noise-free while features may contain noise \cite{ren2013noise,RePCD-Net,3D-pc-denoising}. 
\begin{figure}[htpb]
     \centering
    \includegraphics[width=1\linewidth]{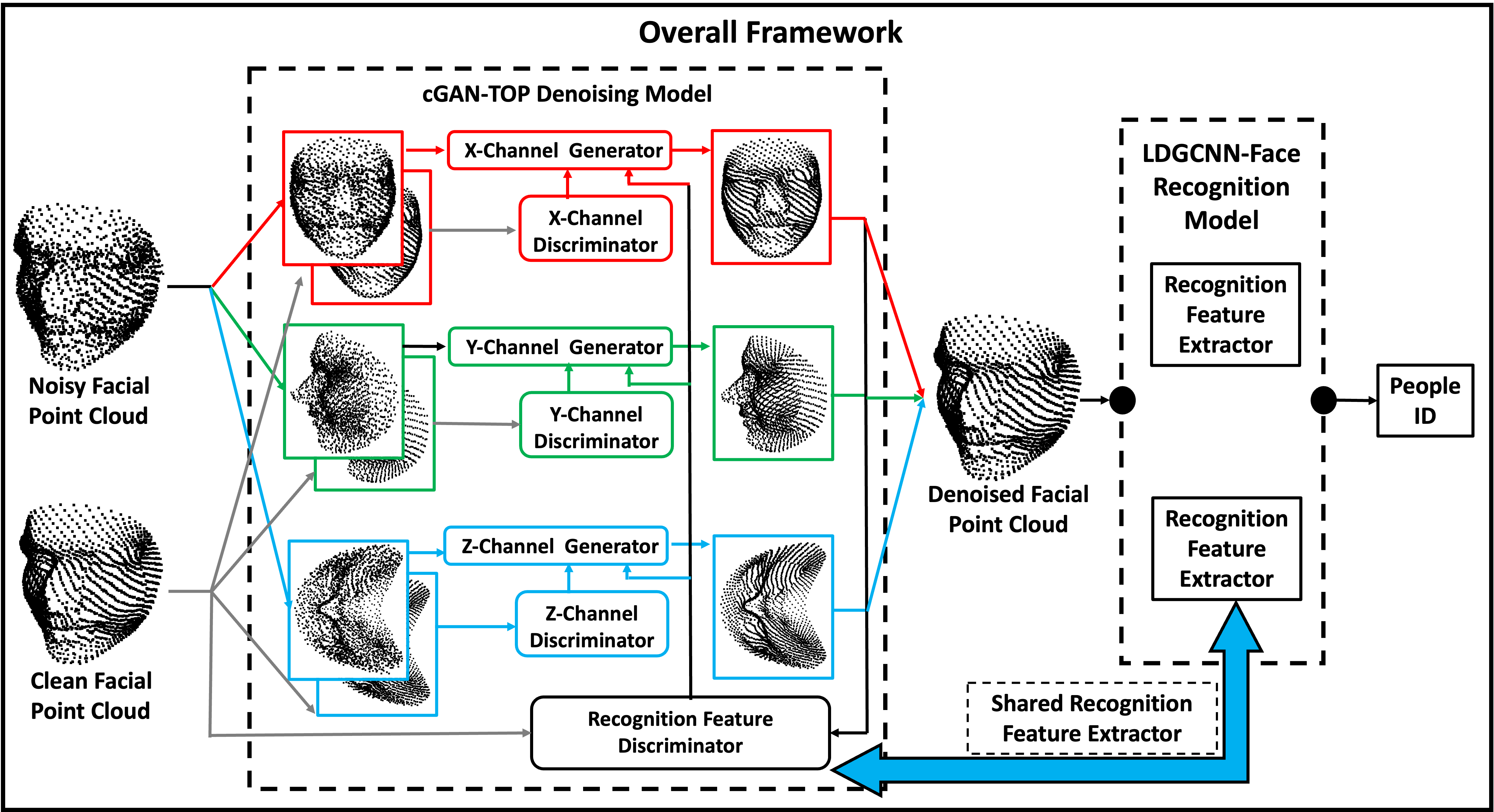} 
     \caption{Overview of the proposed end-to-end framework for 3D face point cloud denoising and recognition.}
     \label{fig:overall-framework}
\end{figure}

To tackle these challenges, the proposed method integrates noise removal and face recognition into one end-to-end framework. As shown in Fig. \ref{fig:overall-framework}, the proposed model consists of a cGAN-TOP for denoising and an LDGCNN for recognition. To remove the noise in the point cloud, its 3D coordinates are mapped into three planes, \ie, $x-y$, $y-z$ and $z-x$ planes. In each of the planes, the point coordinate orthogonal to the plane is treated as the gray value, while the other two remain as the point spatial coordinates in the plane. The 3D point cloud is hence mapped into images of three orthogonal planes. Then a cGAN \cite{yang2022rain} is designed for image denoising from each of the planes. The denoised gray values in the three planes are then combined to form the denoised point cloud. We then adapt a latest general point cloud object  recognition model LDGCNN \cite{arxiv/LDGCNN} for 3D face recognition, which comprehensively extracts both global and local features using a Graph Neural Network. 

\subsection{cGAN-TOP denoising Model}
\begin{figure*}[htbp]
     \centering
     \includegraphics[width=0.9\linewidth]{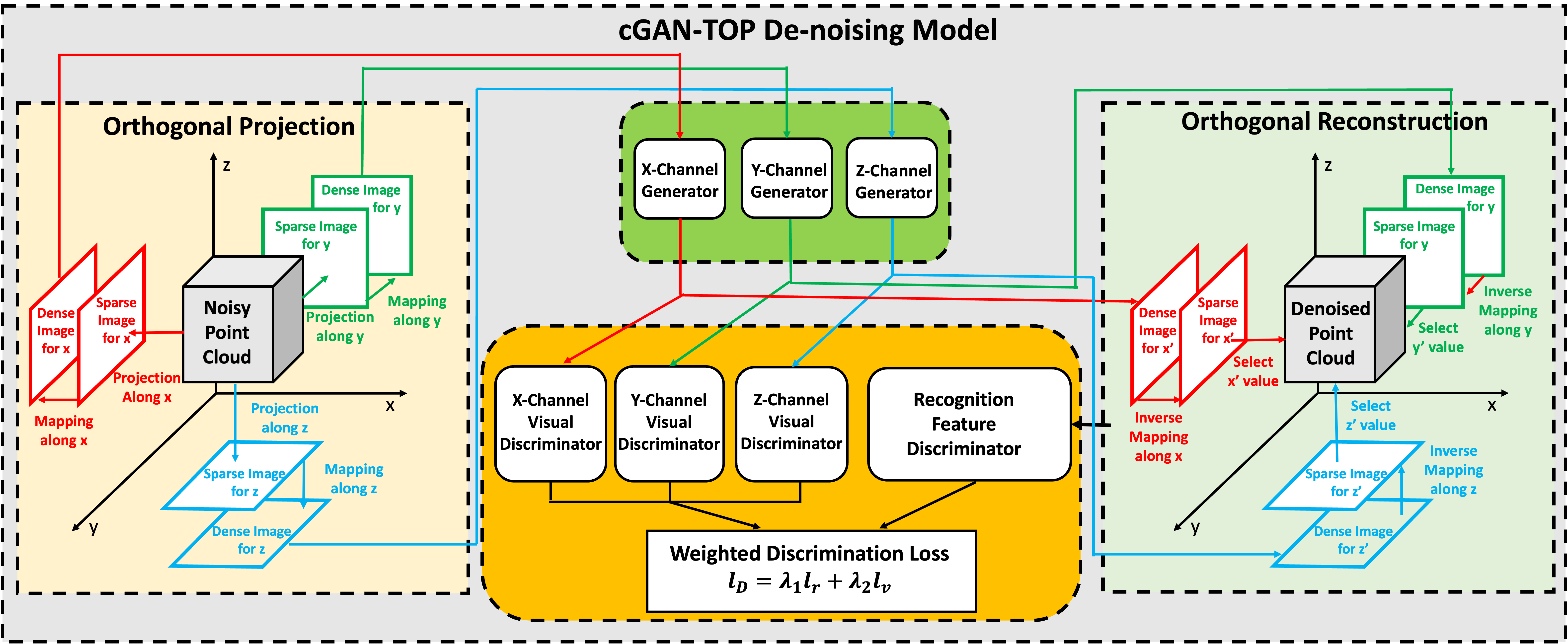} 
     \caption{Block diagram of cGAN-TOP for denoising. The point cloud is first projected into three orthogonal planes with the coordinate orthogonal to the plane as its gray value, and then processed by a cGAN in each plane. Three generators and discriminators are designed to generate denoised images. Each discriminator evaluates the visual cleanness of $x, y, z$ value respectively. The denoised point cloud is then reconstructed from these three planes. The Recognition Feature Discriminator evaluates the discriminant power of the generated point cloud for recognition.}
     \label{fig:CGAN-structure}
 \end{figure*}

The proposed cGAN-TOP is shown in Fig.~\ref{fig:CGAN-structure}, which consists of three sets of generators and discriminators, one set for each of the three orthogonal planes. The motivations behind  this design are two-fold: 1) Traditional 2D and 3D CNNs \cite{zhang2022spatial, chen2022attention, he2023hierarchical} can't directly handle sparse data structures like point clouds. 2) 3D Graph Convolutional Networks can't effectively process noisy 3D face models due to the thin surface-like characteristics of face point clouds. Therefore, we apply three orthogonal projections to decouple $(x,y,z)$ values of a point cloud into three orthogonal planes and take the coordinate orthogonal to the plane as the gray value of the point in the 2D plane. Take projection along z-axis as an example, the 3D point cloud is first projected as a sparse image, where each pixel is represented by a gray value $z$ with an associated coordinate $(x,y)$. We then map the coordinates of this sparse image into a structured image, \ie,  $f_z:(x,y)\rightarrow(x',y')$. After processing by cGAN, the three denoised images are mapped back to the sparse images with the inverse function $f'$, and the denoised coordinates are combined back to a point cloud.

 \begin{figure}[htpb]
     \centering     \includegraphics[width=\linewidth]
     {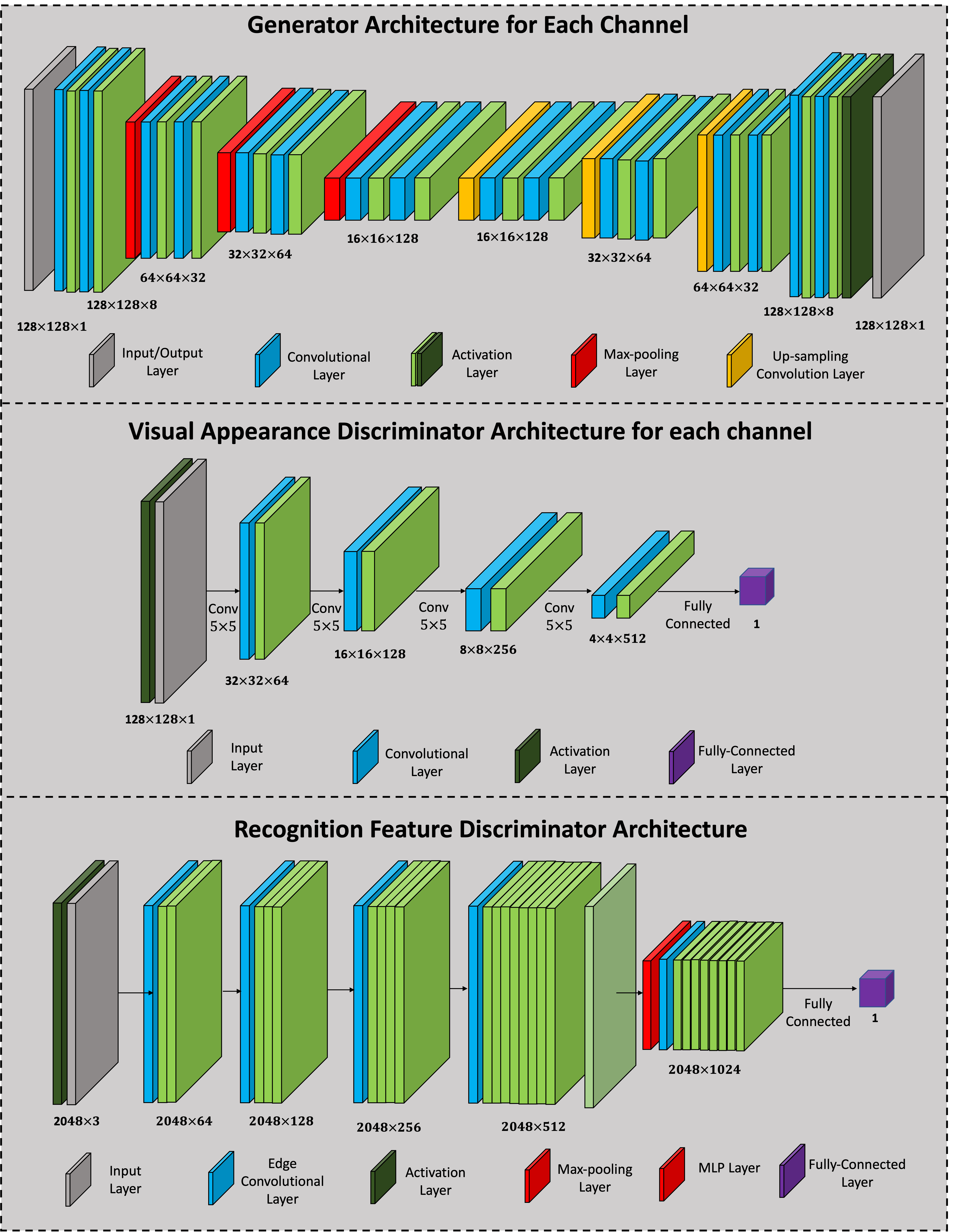} 
     \caption{The structure of Generator and Visual Appearance Discriminator of each channel, and Recognition Feature Discriminator in cGAN-TOP.}
     \label{fig:Generator-Discriminator-structure}
 \end{figure}
 
The proposed cGAN-TOP contains three generators, one for each plane as shown in Fig.~\ref{fig:Generator-Discriminator-structure}. Each generator is a UNet-like network \cite{MICCAI15/U-Net} consisting of an encoder network and a decoder network. The encoder follows the general convolutional neural network \cite{zhang2022spatial} with repeated convolution-pooling units. The decoder uses symmetric processing with upsampling or de-convolution layers. The convolution kernels are of size $5\times5$ and each convolution is followed by a rectified linear unit (ReLU) layer and a max-pooling layer for downsampling. The final image is generated using a convolution layer. 

The discriminator network is designed to achieve the following. 1) The denoised point cloud should enable accurate face recognition. 2) The denoised point cloud should be visually and geometrically clean. Hence, two kinds of discriminators are designed. Each of the three Visual Appearance Discriminator (VAD) is designed similar to DCGAN \cite{ICLR16/DCGAN} as shown in Fig.~\ref{fig:Generator-Discriminator-structure}, which evaluates the visual cleanness of the generated point cloud from the generator using convolution operations and returns the loss $l_{v}$ for the VAD. Moreover, a Recognition Feature Discriminator (RFD) shown in Fig.~\ref{fig:Generator-Discriminator-structure} is designed to evaluate the similarity of the linked features with the clean clouds with a loss of $l_{r}$, to ensure that the denoised point cloud is suitable for face recognition. The final loss $l_D$ for the discriminator is weighted by hyper-parameters $\lambda_1$ and $\lambda_2$ as $l_D =\lambda_1 l_{r} + \lambda_2 l_{v}$, which are tuned as $\lambda_1=0.67$ and $\lambda_2=0.33$ to achieve the optimal performance.

\subsection{LDGCNN Face Recognition Model}
\begin{figure}[htpb]
     \centering
     \includegraphics[width=1\linewidth]{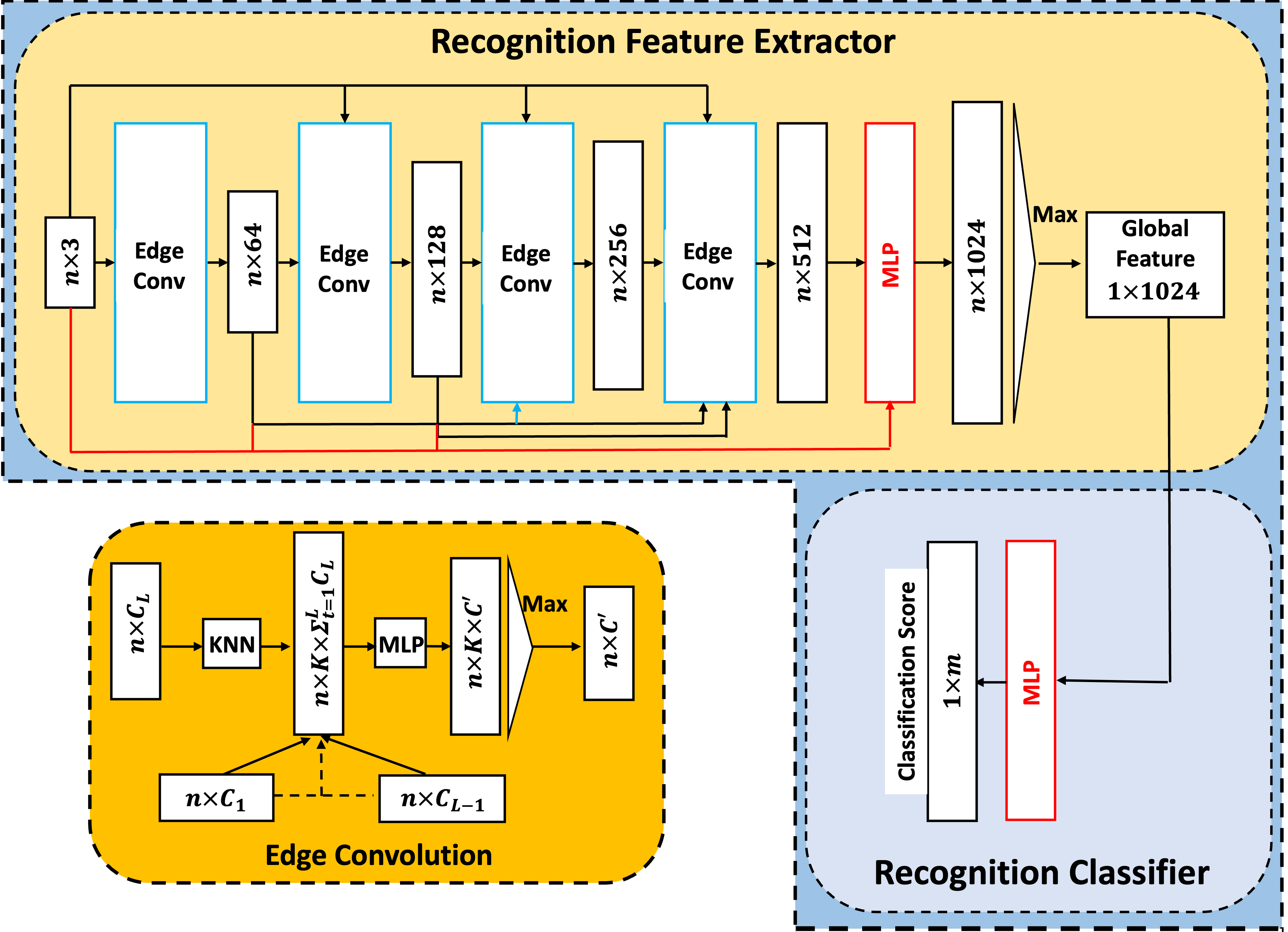} 
     \caption{LDGCNN face recognition model.}
     \label{fig:LDGCNN-structure}
 \end{figure}

An LDGCNN model \cite{arxiv/LDGCNN} is adapted for face recognition, which hierarchically links both the local features and neighboring features of multiple scales in an elegant and high performance manner. As shown in Fig.~\ref{fig:LDGCNN-structure}, the Edge Convolution applies KNN and multi-layer perception with sharing parameters to extract the local and global features of different scales. The shortcuts between different layers are added to link the hierarchical features to derive the informative edge vectors. Slightly different from the original design, we modify the number of channels in a gradually increasing manner to better explore the features.

%-------------------- Experiments --------------------%
\section{Experimental Results}
\label{sec:experiments}

\subsection{Experimental Settings}
\begin{figure}[ht]
     \centering
     \includegraphics[width=1\linewidth]{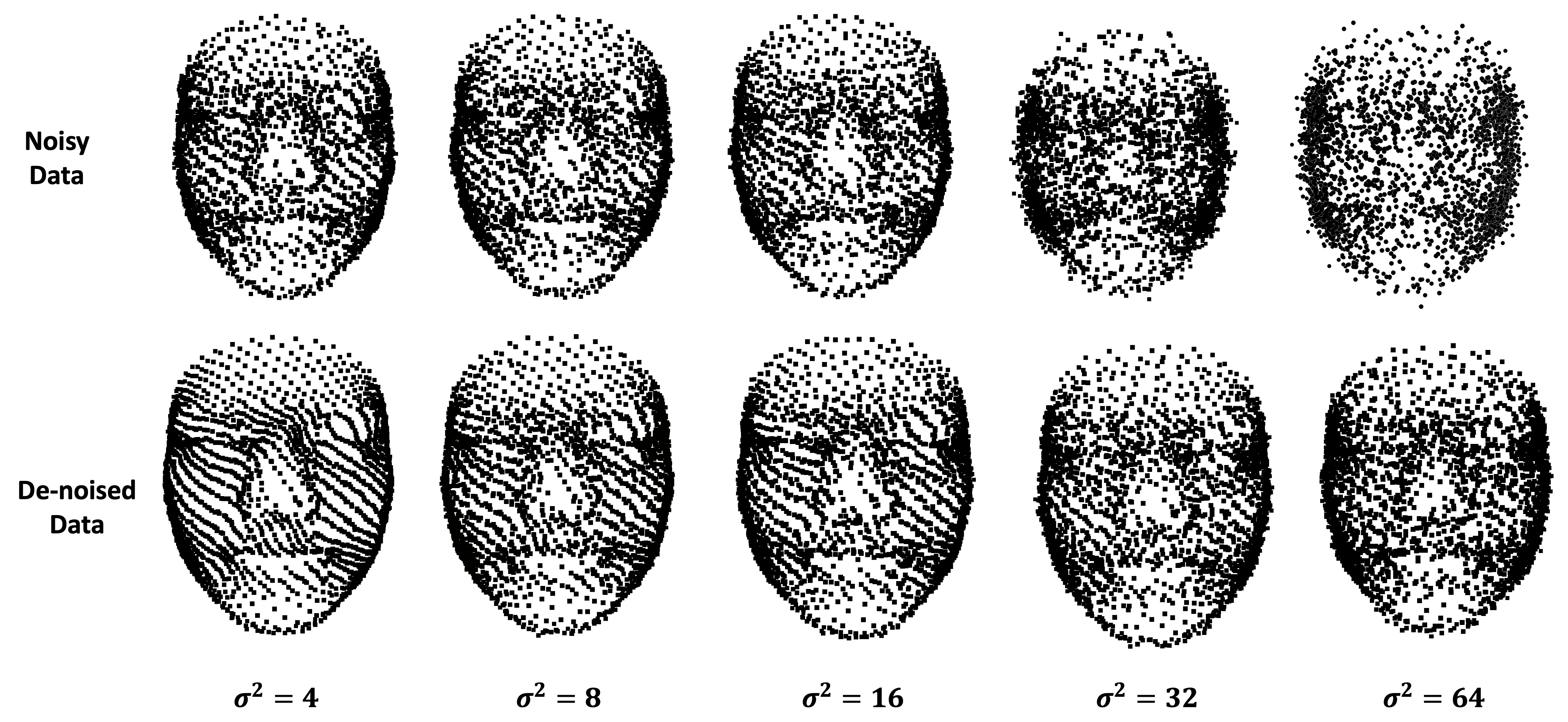} 
     \caption{Sample noisy point clouds and denoised point clouds for $\sigma^2 = \{4,8,16,32,64\}$, respectively.}
     \label{fig:noise_comparison}
 \end{figure}

The proposed method is evaluated on the Bosphorus dataset \cite{Bosphorus}, a commonly used 3D face point cloud dataset. It contains 45,000 3D faces of 150 individuals, including variations in expression, occlusion, and pose. We use two common settings \cite{AAAI21/Learn-low-quality}.
1) \textbf{Neutral Setting}: Faces with a neutral expression ($60\%$ of the dataset) are selected as the training data and the rest are test data.
2) \textbf{Random Setting}: The training data are randomly sampled from all expressions ($60\%$ of the dataset) and the rest for testing.  
We inject 5 different levels of Gaussian noise into point clouds with $\sigma^2 = \{4, 8, 16, 32, 64\}$, following the same experimental settings used in \cite{CVPR19/Led3D} and \cite{AAAI21/Learn-low-quality}.
Some sample noisy point clouds are shown in Fig.~\ref{fig:noise_comparison}. 

\subsection{Comparisons with State-of-the-art Methods}

We first compare the proposed method with state-of-the-art 3D face recognition models such as Led3D \cite{CVPR19/Led3D} and FER \cite{AAAI21/Learn-low-quality}. Table \ref{tab:recognition-accuracy} shows the average recognition accuracy on noisy point clouds under both \textbf{Neutral Setting} and \textbf{Random Setting}. For all noise levels, the proposed model significantly outperforms the previous methods, with a maximum improvement of $14.81\%$ when $\sigma^2=64$ under the \textbf{Random Setting}.

\begin{table}[ht]
\setlength\tabcolsep{1pt}
\caption{Comparison of recognition rate on noisy point cloud.}
\centering
\scalebox{0.95}{
    \begin{tabular}{c|ccc|ccc}
    \bottomrule
        &               &           Neutral Setting&   &                         &          Random Setting      &\\ 
        $\sigma^2$&     LeD3D&          FER&            Ours&                   LeD3D&          FER&            Ours \\\hline
        4&              95.08&           96.88&         \textbf{98.74}&          90.07&         94.87&          \textbf{96.13}\\
        8&              92.87&           95.42&         \textbf{97.78}&          84.33&         90.85&          \textbf{95.53}\\
        16&             87.54&           91.72&         \textbf{93.44}&          72.77&         81.09&          \textbf{88.33}\\ 
        32&             74.66&           82.33&         \textbf{87.57}&          50.51&         61.32&          \textbf{71.26}\\
        64&             55.37&           60.37&         \textbf{72.04}&          28.73&         36.80&          \textbf{51.61} \\\bottomrule
    \end{tabular}
}
\label{tab:recognition-accuracy}
\end{table}

Table \ref{tab:visual-cleanness} shows the comparisons of visual cleanness with state-of-the-art denoising models, PCNet \cite{PCNet}, DMR \cite{DMR}, and SB \cite{score-based}, where Chamfer distance (CD) and Point-to-Mesh distance (P2M) are used as evaluation metrics. The proposed method shows advantages at a high noise level and comparable performance to SB \cite{score-based} at a low noise level.
\begin{table}[htpb]
\caption{Comparisons of visual cleanness in terms of Chamfer distance (CD) and Point-to-Mesh distance (P2M).} % in terms of CD and P2M.}
\centering
\scalebox{0.9}{
    \begin{tabular}{c|c|ccccccc}
    \bottomrule
            $\sigma^2$&     Metric&     PCNet \cite{PCNet} &      DMR \cite{DMR} &        SB \cite{score-based}&         Ours&  \\\hline
    \multirow{2}{*}{4}&     CD&         3.73&       4.64&       \textbf{2.87}&       2.89\\
                    &       P2M&        1.49&       1.83&       0.46&       \textbf{0.48}\\\hline
    \multirow{2}{*}{8}&     CD&         7.86&       5.12&       3.87&       \textbf{3.82}\\
                    &       P2M&        4.16&       2.35&       1.23&       \textbf{1.18}\\\hline
    \multirow{2}{*}{16}&    CD&         13.81&      6.94&       5.02&      \textbf{4.83}\\ 
                    &       P2M&        6.19&       3.27&       2.14&      \textbf{1.86}\\\hline
    \multirow{2}{*}{32}&    CD&         24.93&      14.56&      8.75&      \textbf{6.83}\\ 
                    &       P2M&        12.37&      7.25&       4.28&      \textbf{3.24}\\\hline
    \multirow{2}{*}{64}&    CD&         42.87&      24.91&      18.55&      \textbf{13.26}\\ 
                    &       P2M&        20.35&      12.11&      9.67&      \textbf{6.28}\\\bottomrule
    \end{tabular}
}
\label{tab:visual-cleanness}
\end{table}

To better understand the source of improvements, we conduct an ablation study on the two discriminators of the proposed cGAN-TOP: 1) Use Recognition Feature Discriminator only; 2) Use Visual Feature Discriminator only; 3) Use both discriminators (Full). 
The results shown in Table \ref{tab:ablation-study} indicate that: 1) Use Visual Feature Discriminator only significantly reduces the recognition accuracy as there lacks a supervision for recognition in this case; and 2) Use Recognition Feature Discriminator only also reduce the recognition accuracy since purely targeting at recognition ignores the noise in the point cloud. Thus, we conclude that the proposed method could simultaneously achieve high recognition accuracy on noisy point cloud and produce visually clean point clouds.

\begin{table}[hh]
\setlength\tabcolsep{0.75pt}
\caption{Ablation studies of cGAN-TOP using different discriminators in terms of recognition rate.}
\scalebox{1}{
\centering
    \begin{tabular}{c|ccc|ccc}
    \bottomrule
        &               &            Neutral Setting&   &           &          Random Setting      &\\ 
        $\sigma^2$&     VFD &      LRFD &        Full&               VFD&      LRFD &        Full \\\hline
        4&              86.76&           92.88&         \textbf{98.74}&          84.32&         90.71&          \textbf{96.13}\\
        8&              78.63&           90.42&         \textbf{97.78}&          72.43&         87.29&          \textbf{95.53}\\
        16&             72.67&           86.72&         \textbf{93.44}&          69.72&         84.34&          \textbf{88.33}\\ 
        32&             64.21&           82.33&         \textbf{87.57}&          62.39&         70.01&          \textbf{71.26}\\
        64&             53.34&           62.37&         \textbf{72.04}&          46.51&         49.42&          \textbf{51.61} \\\bottomrule
    \end{tabular}
}
\label{tab:ablation-study}
\end{table}

%-------------------- Conclusions --------------------%
\section{Conclusion}
In this paper, we propose an end-to-end 3D face point cloud denoising and recognition model to provide both high recognition accuracy and visually clean point clouds. The proposed cGAN-TOP provides an efficient and effective way to directly remove the noise in the 3D coordinates of the point cloud by using cGAN on each of the three orthogonal planes. On each plan, the proposed cGAN-TOP strategically treats one of the coordinates as the feature and denoises it using cGAN. Two discriminators are designed to ensure both the high recognition rate and cleanness of the point cloud. An LDGCNN is adapted for 3D face recognition. The proposed method is compared with state-of-the-art methods on the Bosphorus dataset. The proposed method significantly outperforms the compared methods in most of the settings. 

\vfill\clearpage

\small
\bibliographystyle{IEEEbib}
\bibliography{main-arxiv}

\begin{thebibliography}{10}

\bibitem{CVPR18/Learn-Large-scale-3D-Face}
S.~Gilani and A.~Mian,
\newblock ``Learning from millions of {3D} scans for large-scale {3D} face
  recognition,''
\newblock in {\em {IEEE} Conference on Computer Vision and Pattern
  Recognition}, 2018, pp. 1896--1905.

\bibitem{IJCB21/PointFace}
C.~Jiang, S.~Lin, W.~Chen, F.~Liu, and L.~Shen,
\newblock ``{PointFace}: Point set based feature learning for {3D} face
  recognition,''
\newblock in {\em International {IEEE} Joint Conference on Biometrics}, 2021,
  pp. 1--8.

\bibitem{FeatureBoost-3D}
J.~Liu, H.~Ding, A.~Shahroudy, L.~Duan, X.~Jiang, G.~Wang, and A.~Kot,
\newblock ``Feature boosting network for {3D} pose estimation,''
\newblock {\em IEEE transactions on pattern analysis and machine intelligence},
  vol. 42, no. 2, pp. 494--501, 2019.

\bibitem{Vessel-seg}
X.~Wang, X.~Jiang, and J.~Ren,
\newblock ``Blood vessel segmentation from fundus image by a cascade
  classification framework,''
\newblock {\em Pattern Recognition}, vol. 88, pp. 331--341, 2019.

\bibitem{HCI22/Face2Statistics}
Z.~Xiong, J.~Wang, W.~Jin, J.~Liu, Y.~Duan, Z.~Song, and X.~Peng,
\newblock ``Face2statistics: user-friendly, low-cost and effective alternative
  to in-vehicle sensors/monitors for drivers,''
\newblock in {\em {International Conference on Human-Computer Interaction}},
  2022.

\bibitem{arxiv/LDGCNN}
K.~Zhang, M.~Hao, J.~Wang, C.~W. de~Silva, and C.~Fu,
\newblock ``Linked dynamic graph {CNN:} learning on point cloud via linking
  hierarchical features,''
\newblock {\em CoRR}, vol. abs/1904.10014, 2019.

\bibitem{Hand-craft-1}
S.~J. Belongie, J.~Malik, and J.~Puzicha,
\newblock ``Shape matching and object recognition using shape contexts,''
\newblock {\em {IEEE} Trans. Pattern Anal. Mach. Intell.}, vol. 24, no. 4, pp.
  509--522, 2002.

\bibitem{2D-based-1}
Z.~Wu, S.~Song, A.~Khosla, F.~Yu, L.~Zhang, X.~Tang, and J.~Xiao,
\newblock ``{3D ShapeNets}: {A} deep representation for volumetric shapes,''
\newblock in {\em {IEEE} Conference on Computer Vision and Pattern
  Recognition}, 2015, pp. 1912--1920.

\bibitem{2D-based-2}
M.~Yavartanoo, E.~Kim, and K.~M. Lee,
\newblock ``{SPNet}: Deep {3D} object classification and retrieval using
  stereographic projection,''
\newblock in {\em Asian Conference on Computer Vision}, 2018, pp. 691--706.

\bibitem{2D-based-3}
S.~Chen, L.~Zheng, Y.~Zhang, Z.~Sun, and K.~Xu,
\newblock ``{VERAM:} view-enhanced recurrent attention model for {3D} shape
  classification,''
\newblock {\em {IEEE} Trans. Vis. Comput. Graph.}, vol. 25, no. 12, pp.
  3244--3257, 2019.

\bibitem{arxiv/DGCNN}
Y.~Wang, Y.~Sun, Z.~Liu, S.~E. Sarma, M.~M. Bronstein, and J.~M. Solomon,
\newblock ``Dynamic graph {CNN} for learning on point clouds,''
\newblock {\em CoRR}, vol. abs/1801.07829, 2018.

\bibitem{IJCNN16/PointNet}
A.~Garcia{-}Garcia, F.~Gomez{-}Donoso, J.~G. Rodr{\'{\i}}guez,
  S.~Orts{-}Escolano, M.~Cazorla, and J.~Azor{\'{\i}}n L{\'{o}}pez,
\newblock ``{PointNet}: {A 3D} convolutional neural network for real-time
  object class recognition,''
\newblock in {\em International Joint Conference on Neural Networks}, 2016, pp.
  1578--1584.

\bibitem{3D-based}
Z.~Zhang, F.~Da, and Y.~Yu,
\newblock ``Learning directly from synthetic point clouds for "in-the-wild"
  {3D} face recognition,''
\newblock {\em Pattern Recognit.}, vol. 123, pp. 108394, 2022.

\bibitem{NIPS17/PointNet++}
C.~R. Qi, L.~Yi, H.~Su, and L.~J. Guibas,
\newblock ``Pointnet++: Deep hierarchical feature learning on point sets in a
  metric space,''
\newblock in {\em Advances in Neural Information Processing Systems}, 2017, pp.
  5099--5108.

\bibitem{AAAI21/Learn-low-quality}
Z.~Zhang, C.~Yu, S.~Xu, and H.~Li,
\newblock ``Learning flexibly distributional representation for low-quality
  {3D} face recognition,''
\newblock in {\em {AAAI} Conference on Artificial Intelligence}, 2021, pp.
  3465--3473.

\bibitem{CVPR19/Led3D}
G.~Mu, D.~Huang, G.~Hu, J.~Sun, and Y.~Wang,
\newblock ``{Led3D}: {A} lightweight and efficient deep approach to recognizing
  low-quality {3D} faces,''
\newblock in {\em {IEEE} Conference on Computer Vision and Pattern
  Recognition}, 2019, pp. 5773--5782.

\bibitem{CVPR19/3D-Point-Set-Upsampling}
Y.~Wang, S.~Wu, H.~Huang, D.~Cohen{-}Or, and O.~Sorkine{-}Hornung,
\newblock ``Patch-based progressive {3D} point set upsampling,''
\newblock in {\em {IEEE} Conference on Computer Vision and Pattern
  Recognition}, 2019, pp. 5958--5967.

\bibitem{yang2022rain}
F.~Yang, J.~Ren, Z.~Lu, J.~Zhang, and Q.~Zhang,
\newblock ``Rain-component-aware capsule-{GAN} for single image de-raining,''
\newblock {\em Pattern Recognition}, vol. 123, pp. 108377, 2022.

\bibitem{Bosphorus}
A.~Savran, N.~Aly{\"{u}}z, H.~Dibeklioglu, O.~{\c{C}}eliktutan,
  B.~G{\"{o}}kberk, B.~Sankur, and L.~Akarun,
\newblock ``Bosphorus database for {3D} face analysis,''
\newblock in {\em Biometrics and Identity Management}, 2008, vol. 5372, pp.
  47--56.

\bibitem{ren2013noise}
J.~Ren, X.~Jiang, and J.~Yuan,
\newblock ``Noise-resistant local binary pattern with an embedded
  error-correction mechanism,''
\newblock {\em IEEE Transactions on Image Processing}, vol. 22, no. 10, pp.
  4049--4060, 2013.

\bibitem{RePCD-Net}
H.~Chen, Z.~Wei, X.~Li, Y.~Xu, M.~Wei, and J.~Wang,
\newblock ``Repcd-net: Feature-aware recurrent point cloud denoising network,''
\newblock {\em International Journal of Computer Vision}, vol. 130, no. 3, pp.
  615--629, 2022.

\bibitem{3D-pc-denoising}
C.~Shi, C.~Wang, X.~Liu, S.~Sun, B.~Xiao, X.~Li, and G.~Li,
\newblock ``Three-dimensional point cloud denoising via a gravitational feature
  function,''
\newblock {\em Applied Optics}, vol. 61, no. 6, pp. 1331--1343, 2022.

\bibitem{zhang2022spatial}
J.~Zhang, Q.~Zhang, J.~Ren, Y.~Zhao, and J.~Liu,
\newblock ``Spatial-context-aware deep neural network for multi-class image
  classification,''
\newblock in {\em IEEE International Conference on Acoustics, Speech and Signal
  Processing}, 2022, pp. 1960--1964.

\bibitem{chen2022attention}
S.~Chen, W.~He, J.~Ren, and X.~Jiang,
\newblock ``Attention-based dual-stream vision transformer for radar gait
  recognition,''
\newblock in {\em IEEE International Conference on Acoustics, Speech and Signal
  Processing}, 2022, pp. 3668--3672.

\bibitem{he2023hierarchical}
W.~He, J.~Zhang, J.~Ren, R.~Bai, and X.~Jiang,
\newblock ``Hierarchical convit with attention-based relational reasoner for
  visual analogical reasoning,''
\newblock in {\em {AAAI} Conference on Artificial Intelligence}, 2023.

\bibitem{MICCAI15/U-Net}
O.~Ronneberger, P.~Fischer, and T.~Brox,
\newblock ``U-net: Convolutional networks for biomedical image segmentation,''
\newblock in {\em Medical Image Computing and Computer-Assisted Intervention},
  2015, vol. 9351, pp. 234--241.

\bibitem{ICLR16/DCGAN}
A.~Radford, L.~Metz, and S.~Chintala,
\newblock ``Unsupervised representation learning with deep convolutional
  generative adversarial networks,''
\newblock {\em arXiv preprint arXiv:1511.06434}, 2015.

\bibitem{PCNet}
M.~Rakotosaona, V.~La~Barbera, P.~Guerrero, N.~J. Mitra, and M.~Ovsjanikov,
\newblock ``Pointcleannet: Learning to denoise and remove outliers from dense
  point clouds,''
\newblock in {\em Computer Graphics Forum}, 2020, vol.~39, pp. 185--203.

\bibitem{DMR}
S.~Luo and W.~Hu,
\newblock ``Differentiable manifold reconstruction for point cloud denoising,''
\newblock in {\em ACM international conference on multimedia}, 2020, pp.
  1330--1338.

\bibitem{score-based}
S.~Luo and W.~Hu,
\newblock ``Score-based point cloud denoising,''
\newblock in {\em IEEE International Conference on Computer Vision}, 2021, pp.
  4583--4592.

\end{thebibliography}

\end{document}